\pdfoutput=1

\documentclass[11pt]{article}

\usepackage{emnlp2021}

\usepackage{times}
\usepackage{latexsym}

\usepackage{graphicx}

\usepackage[namelimits]{amsmath} 
\usepackage{amssymb}             
\usepackage{amsfonts}            
\usepackage{mathrsfs}
\usepackage{color}   
\usepackage{CJKutf8}   
\usepackage[T1]{fontenc}

\usepackage[utf8]{inputenc}

\usepackage{microtype}

\title{Context-Adaptive Document-Level Neural Machine Translation}

\author{Linlin Zhang \\
  Zhejiang University / China \\
  \texttt{11921133@zju.edu.cn}}

\begin{document}
\maketitle

\begin{CJK}{UTF8}{gkai}  
\begin{abstract}
Most existing document-level neural machine translation (NMT) models leverage a fixed number of the previous or all global source sentences to handle the context-independent problem in standard NMT. 
However, the translating of each source sentence benefits from various sizes of context, and inappropriate context may harm the translation performance. This work introduces a data-adaptive method that enables the model to adopt the necessary and helpful context. Specifically, we introduce a light predictor into two document-level translation models to select the explicit context. 
Experiments demonstrate the proposed approach can significantly improve the performance over the previous methods with a gain up to 1.99 BLEU points.
\end{abstract}

\section{Introduction}

Neural machine translation (NMT) based on the encoder-decoder framework has advanced translation performance in recent years \cite{DBLP:conf/nips/SutskeverVL14,DBLP:journals/corr/BahdanauCB14,Wu2016GooglesNM,Hassan2018AchievingHP}.
Instead of translating sentences in isolation, 
document-level machine translation (DocMT) methods are proposed to capture discourse dependencies across sentences by considering a document as a whole. 

Current DocMT systems usually leverage a fixed amount of source or target context sentences while translating~\citep{voita2018context,zhang2018improving,werlen2018document,yang2019enhancing,voita2019good,zhu2019incorporating,mansimov2020capturing,xu2020efficient}. 
It is observed in \citet{kang2020dynamic}, which evaluates some previous DocMT models using different contexts, and the results shown that less context can get a higher BLEU than a fixed previous context sometimes. Thus, the translation model may need a more flexible context instead of a fixed static context, choosing a proper context becomes vital in DocMT systems \citep{maruf2019selective,kang2020dynamic,saunders2020using}.

To tackle this problem, \citet{maruf2019selective} proposed a selective attention approach that normalizes the attention weights via the sparsemax function instead of the softmax.
In their model, the sparsemax converts the low probability in softmax to zero, only keeping the sentences with high probability. 
However, this method focuses on selective attention weights and cannot handle cases where the source sentence achieves the best translation result without using any context. 

In this paper, we propose a novel framework to predict the most appropriate context for model translation. 
To achieve this goal, we directly utilize a lightweight predictor, which takes encoder outputs as inputs, to predict the probabilities of different context options.
We take their corresponding classification losses as training signals to update this lightweight predictor.
Based on this, the best context for each source sentence can be selected during inference with only introducing little time cost. 
Also, the candidate contexts are limited in the most relevant pre-sentence and post-sentence, without searching from the enormous scope like the previous work.
Experimental results prove that our proposed approach can significantly outperform the previous baselines with a margin up to 1.99 BLEU points.

Our main contributions can be summarized as follows:
\begin{itemize} 
\item  Our method can select the appropriate context by introducing a lightweight predictor. The predictor and the DocMT are trained jointly with a few additional parameters.
\item  Our method is applied to two basic DocMT models where one utilizes source context information, and another uses both source and target. The two models gain significant improvements with our proposed method.
\end{itemize}

\vspace{-12pt}
\begin{figure}
    \centering
    \includegraphics[scale=0.50]{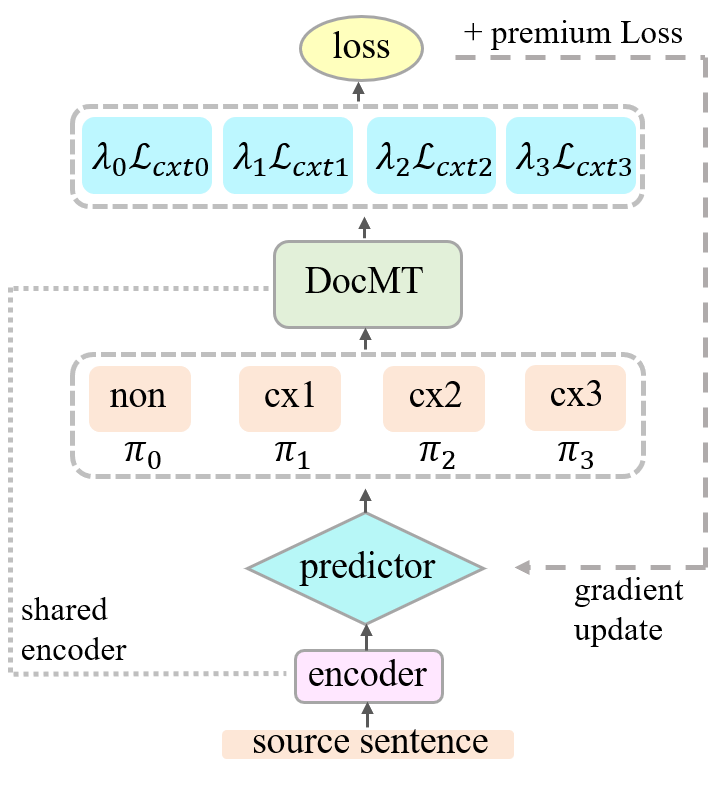}
    \vspace{-8pt}
    \caption{The framework for Context-adaptive DocMT. $\pi$ is the context option probability from the predictor. $\lambda$ and premium losses are detailed in section\ref{sec-training}.}
    \vspace{-7pt}
    \label{fig:my_predictor}
\end{figure}

\section{Context-Adaptive DocMT}

In this section, we introduce our Context-adaptive DocNMT by adding a context predictor.

\subsection{Document-Level Translation}

Compared with sentence-level MT that translates a sentence $X=\left\{x_{1}, \cdots x_{S}\right\}$ into a target sentence $Y=\left\{y_{1}, \cdots y_{T}\right\}$, DocMT takes advantage of the contextual information $C$ of the document. The DocMT model can be trained to minimize the negative log-likelihood loss as:
\vspace{-6pt}
\begin{equation}
    \mathcal{L}_{D M T}=-\sum_{t=1}^{T} \log P\left(y_{t} \mid y<t, X, C ; \theta\right)
\end{equation}
\vspace{-10pt}

\subsection{Context Predictor}

To select the appropriate context, we predict context based on the current inputs. As shown in Figure~\ref{fig:my_predictor}, the source sentence is processed by the encoder of the DocMT model. Then we feed the corresponding encoder output to the predictor, to calculate the probability $\pi_{i}$ of the $i$-th choice.

Considering there are $N$ different options for context to choose, including not adopting any context (empty context). Each source sentence has its own preference for some context-added options. 
We leverage the cross-entropy loss as training signals to learn probability. Therefore, for each context-added option $c x t_{i}$, we calculate the probability $\lambda_{i}$ of $c x t_{i}$, and then weigh the corresponding translation loss $\mathcal{L}_{M T}^{i}$ with the selected context.

\subsubsection{Training Loss}
\label{sec-training} 

For each source sentence $X=\left\{x_{1}, \cdots x_{S}\right\}$, where S is the length of the input, $H=\left\{h_{1}, \cdots h_{S}\right\}$ is the corresponding output of DocMT encoder. The light predictor or classifier leverages the averaged encoder outputs $H_{s}=\frac{1}{s} \sum_{k=1}^{S} h_{\mathrm{k}}$ to predict the possibility of context-selected options. 
Then, uses softmax to calculate the possibility:
\vspace{-3pt}
\begin{equation}
    \pi=\operatorname{softmax}\left(H_{s} \times W+b\right)
\end{equation}
\vspace{-3pt}
where $\mathrm{W} \in \mathbb{R}^{d \times N}$ is the projection weight matrix. The weight $\lambda_{i}$ of $\mathcal{L}_{M T}^{i}$ for each context option $c x t_{i}$:
\vspace{-3pt}
\begin{equation}
  \lambda_{i}=\frac{\exp \left(\left(\log \left(\pi_{i}\right)+g_{i}\right) / \tau\right)}{\sum_{j=1}^{N} \exp \left(\left(\log \left(\pi_{j}\right)+g_{j}\right) / \tau\right)}
\end{equation}
\vspace{-5pt}
where noise $g_{i}$ is sampled from Gumbel distribution, and $\tau$ is a constant temperature. The cross-entropy loss of training is a weighted combination of confidence $\lambda_{i}$ and corresponding loss $\mathcal{L}_{M T}^{i}$, formulated as:
\vspace{-5pt}
\begin{equation}
    \mathcal{L}_{M T}=\sum_{i=1}^{N} \lambda_{i} \mathcal{L}_{M T}^{i}
\end{equation}
\vspace{-7pt}

\paragraph{Premium Losses}

As mentioned above, the predictor is trained in an unsupervised way, which quickly trained to prefer one specific option. To explore the diverse capability of all options, we incorporate the KL-divergence by adding a diversity loss $\mathcal{L}_{div}$:
\vspace{-5pt}
\begin{equation}
 \begin{aligned}
    \mathcal{L}_{d i v} &=KL(\mathbb{U} \| \mathbb{E}[\pi]) \\
     &=-\frac{1}{N} \sum_{i=1}^{N} \log \left(\mathbb{E}\left[\pi_{i}\right]\right)-\log N
  \end{aligned}
\end{equation}
\vspace{-5pt}

But in the inference stage, we hope the predictor can make an unambiguous selection. So we expect the probability to be far away from the uniform distribution $\mathbb{U}$ via adding the $\mathcal{L}_{u n i}$ loss:
\vspace{-5pt}
\begin{equation}
 \begin{aligned}
    \mathcal{L}_{uni} &=-\mathbb{E}\left[KL(\mathbb{U} \| \pi)\right] \\ &=-\mathbb{E}\left[-\frac{1}{N} \sum_{i=1}^{N} \log \pi_{i}-\log N\right]
 \end{aligned}
\end{equation}
\vspace{-5pt}

Inspired by the BERT \citep{devlin2018bert}, we also use segment embedding and mask strategy on our DocMT models. In this way, the final training objective is to minimize the loss as:
\vspace{-5pt}
\begin{equation}
    \mathcal{L}=\mathcal{L}_{M T}+\beta_{1} \mathcal{L}_{d i v}+\beta_{2} \mathcal{L}_{u n i}+\beta_{3} \mathcal{L}_{mask}
\end{equation}
\vspace{-2pt}
where $\mathcal{L}_{mask}$ is the token masked loss of source sentence, $\beta_{1}$, $\beta_{2}$, and $\beta_{3}$ are the hyper-parameters described in appendix materials.

As a result, the DocMT model and the predictor can be trained jointly.

\subsubsection{Inference Prediction}

When inferring, we use the trained predictor to choose the most suitable context option according to the averaged encoder output $H_{s}$ of source $X$:
\vspace{-5pt}
\begin{equation}
    c x t_{n}=\operatorname{argmax}\left(H_{s} \times W+b\right)
\end{equation}
\vspace{-5pt}
In this way, the framework can dynamically select the appropriate context by the predictor. All options use the same DocMT model.

\section{Experiments}

\begin{table*}[tbp]
\centering
\vspace{-7pt}
 \begin{tabular}{lccccc}
   \hline
      & ZH-EH & & EN-DN & &  \\
    & TED & TED & News & Europarl17 & $\Delta|\theta|$ \\
    \hline
        Sentence Transformer~\citep{vaswani2017attention} & 17.56 & 23.10 & 22.40 & 29.40 \\
        Sentence Transformer (our implementation) & 18.38 & 25.08 & 24.32 & 29.98 & 0.0m \\
    \hline
        HAN~\citep{werlen2018document} & 17.9 & 24.58 & 25.03 & 28.60 & 4.8m \\
        SAN~\citep{maruf2019selective} & n/a & 24.42 & 24.84 & 29.75 & 4.2m \\
        QCN~\citep{yang2019enhancing} & n/a & 25.19 & 22.37 & 29.82 & n/a \\
        Flat-Transformer~\citep{ma2020simple} & n/a & 24.87 & 23.55 & 30.09 & 0.0m \\
        MCN~\citep{zheng2020towards} & 19.1 & 25.10 & 24.91 & 30.40 & 4.7m \\
    \hline
        Context-unit (our implementation) & 19.12 & 25.75 & 24.89 & 30.41 & 7.9m \\
        Context-unit + Our predictor & 19.81 & 26.29 & 25.41 & 30.84 & 9.5m \\
        Concatenate (our implementation) & 18.69 & 25.36 & 24.56 & 30.19 & 0.0m \\
        Concatenate + Our predictor & 20.68 & 26.77 & 25.81 & 31.13 & 2.1m   \\
    \hline
\end{tabular}
\vspace{-3pt}
\caption{\label{all-BLEU}
The translation results of the test sets in BLEU score and increments of the number of parameters over
Transformer baseline $(\Delta|\theta|)$, when compared with several baselines.
}
\vspace{-5pt}
\end{table*}

\begin{table}
\centering
\vspace{-5pt}
\begin{tabular}{lccc}
\hline 
& concatenate & num & percentage \\ 
\hline
1 & $\text{non}\|\text{source}\|\text{non}$ & 336 & $14.80\%$ \\
2 & $\text{pre}\|\text{source}\|\text{non}$ & 578 & 25.45\% \\
3 & $\text{non}\|\text{source}\|\text{pos}$ & 322 & 14.18\% \\
4 & $\text{pre}\|\text{source}\|\text{pos}$ & 1035 & 45.57\% \\
\hline
\end{tabular}
\vspace{-3pt}
\caption{\label{table-option-percen}Statistics of predictor's selections of concatenate model on the TED EN-DE test set.
}
\vspace{-10pt}
\end{table}

\begin{table}
\centering
\begin{tabular}{lrr}
\hline 
model & all tokens & all time \\ 
\hline
sentence-level & 48833 & 659.5s \\
concatenate & 148469 & 1962.3s \\
our model & 108622 & 1878.8s \\
\hline
\end{tabular}
\vspace{-3pt}
\caption{\label{table-inference time}Statistics of models' decoding time of batch size one on the TED EN-DE test set(total 2271 sentences).
}
\vspace{-10pt}
\end{table}
\vspace{-5pt}

\subsection{Datasets and Settings}
For a fair comparison with previous works, we conducted experiments on four widely used document-level parallel datasets of two language pairs:
(1) \textbf{TED (ZH-EN/EN-DE)}: The Chinese-English and English-German TED datasets are from IWSLT 2015 and 2017 evaluation campaigns, respectively. For TED ZH-EN, we take dev2010 as the valid set and tst2010-2013 as the test set. (2) For the \textbf{EN-DE} language pair, we directly use the 3 prepared EN-DE corpora extracted by \citet{maruf2019selective}.

For a fair comparison, we use the same model configuration and training settings as \citet{ma2020simple}, and implement our experiments on Fairseq\footnote{This tool can be accessed via \url{https://github.com/pytorch/fairseq}}, detailed in the appendix materials.



\subsection{Two Basic DocMT}
We apply the above predictor on two prevalent and straightforward basic DocMT models with mini changes. All options use one DocMT model.

\subsubsection{Context-Unit Model}
Many previous DocMT models use two encoders \citep{zhang2018improving,werlen2018document,yang2019enhancing}, one is to process the source sentence, and another is for the context. In a standard Transformer, each layer unit is composed of Multi-head Attention and a point-wise feed-forward network (FFN). The output $F_{i}$ of the $i$-th layer can be calculated from the input $X_{i}$ as:
\vspace{-5pt}
\begin{equation}
    S_{i}^{src}=\operatorname{SelfAttn}_{i}^{src}\left(X_{i}\right)+X_{i}
\end{equation}
\vspace{-15pt}
\begin{equation}
    F_{i}^{src}\left(X_{i}\right)=\mathrm{FFN}_{i}^{src}\left(S_{i}\right)+S_{i}
\end{equation}
\vspace{-5pt}
We add a context-unit to process the context, the context output of $i$-th layer as:
\vspace{-5pt}
\begin{equation}
   S_{i}^{cxt}=\operatorname{SelfAttn}_{i}^{cxt}\left(C_{i}\right)+C_{i}
\end{equation}
\vspace{-20pt}
\begin{equation}
   F_{i}^{cxt}\left(C_{i}\right)=\mathrm{FFN}_{i}^{cxt}\left(S_{i}^{cxt}\right)+S_{i}^{cxt}
\end{equation}
\vspace{-5pt}
Then add the context-unit output with a cross-attention weighted parameter $\alpha$. The final output of $i$ layer as:
\vspace{-5pt}
\begin{equation}
   F_{i}\left(X_{i}, C_{i}\right)=F_{i}^{src}\left(X_{i}\right)+\alpha \operatorname{CrossAttn}_{i}\left(F_{i}^{cxt}\left(C_{i}\right)\right)
\end{equation}
In this model, there are $\mathrm{N}=3$ different context inputs: previous sentence, next sentence, empty context replaced by the source sentence. The values of $\alpha$ and last layer's $CrossAttn$ correspond to different options, while other parameters are the same.

\subsubsection{Concatenate Model}
Concatenating the context and current sentence is a native DocMT model. There are $\mathrm{N}=4$ context-added options, shown in Table~\ref{table-option-percen}

The target sentence concatenation is the same as the source. Inspired by \citet{li2020deep}, we assume that the concatenated input of different lengths corresponds to a different number of model layers. We increase the number of encoder layers by one. In the decoder, corresponding to the above four options in respectively: reduce two layers (exit before the last two layers), reduce one layer, reduce one layer, not reduce. All options work on one DocMT model.

\subsection{Results}

We list the results of experiments
in Table~\ref{all-BLEU}, comparing with a standard sentence-level Transformer and six previous DocMT baselines. Our method is at the lower part.

As shown in Table~\ref{all-BLEU}, our proposed method on the context-unit model and the concatenate model both achieved leading results over other DocMT baselines. With our predictor, the performance of two DocMT models has been significantly improved. For the concatenate model, our method receives $2.30$, $1.69$, $1.49$, $1.15$ BLEU~\citep{papineni2002bleu} gains over the sentence-level Transformer, receives $1.99$, $1.41$, $1.25$, $0.94$ over the concatenate baseline, on TED ZH-EN, TED EN-DE, News and Europarl datasets, respectively.

DocMT models can be trained in two stages: first, train a sentence-level base model, then finetune from the pre-trained model with document-level data. All our context-adaptive DocMT models adopt two stages training to save training time, as in previous works. Due to the $N$ different options, similarly, the same model is trained by $N$ times. Thus, the training time of every epoch increases, but the number of convergence rounds is reduced. As in Table\ref{table-inference time}, when inference, the prediction of the context is increased, and the total decoding time increases very little.
The last column of Table~\ref{all-BLEU} shows that although the training time increases, the model parameters increase very little. Thus the inference speed is controllable. It indicates that the predictor is indeed light-weight.

In Table~\ref{table-option-percen}, there are 4 context options of concatenate model ("non" indicates empty context sentence). $14.80\%$ of the source sentences choose not to use any context, indicating most translations need contextual information. $45.57\%$, less than half prefer to use the context of both the previous and next sentences. As observed from the translation results, contextual information can increase consistency, such as a unified tense, supplementary pronouns, and conjunctions. Nevertheless, as the qualitative example in appendix materials, the DocMT model prefers the previous sentence's tense. In contrast, our model selected the tense of the latter sentence instead of unified all sentences into one tense. The experimental results prove our conjecture that the DocMT model may need a more flexible context instead of a fixed static context.

\subsection{Ablation Study}

We conduct an ablation study by removing
the components of our method from our concatenate model on the TED EN-DE dataset. As shown in Table~\ref{table-ablation}, when we remove the $\mathcal{L}_{uni}$, the gain of the model dropped by 0.14. After removing the $\mathcal{L}_{div}$, the gain dropped by 0.41. Further removing minor DocNMT tips, including segment and adaptive depth, the gain dropped by 0.43. The results show that context predictor helps to improve the translation.

\begin{table}[t]
\centering
\begin{tabular}{lc}
\hline
& BLEU \\
\hline
Our predictor & 26.77 \\
w/o $\mathcal{L}_{uni}$ & 26.63 \\
w/o $\mathcal{L}_{div}$ & 26.22 \\
w/o Doc tips & 25.79 \\
\hline
\end{tabular}
\vspace{-3pt}
\caption{\label{table-ablation} 
Ablation study on the TED dataset.
}
\vspace{-10pt}
\end{table}

\section{Conclusion}
\label{sec:supplementary}
In this paper, we proposed a data-driven framework on DocMT for adaptive context. The method introduces a lightweight predictor to select the most appropriate context without increasing many parameters. Moreover, it is not limited by the specific circumstances of different contexts: empty context, source context, or target context. Experimental results show that the proposed DocMT framework can achieve significant improvements on two baseline models and various datasets.

\bibliography{anthology,custom}
\bibliographystyle{acl_natbib}

\appendix

\section{Example Appendix}
\label{sec:appendix}

\begin{table*} [h]
\begin{tabular}{lp{36.65em}}
\hline
  \text{src+cxt}  &   \textcolor[rgb]{0.40,0.40,0.40}{通过演奏音乐，谈论音乐，这个人从一个偏执，不安的，刚才还在洛杉矶大街上晃悠的流浪汉，变成了一个迷人，博学，优秀的受过朱丽亚音乐学院教育的音乐家。}\textcolor[rgb]{0,0,0}{音乐是良药，音乐改变着我们。} \textcolor[rgb]{0.40,0.40,0.40}{对nathaniel来说，音乐是帮助他开启心智。} \\ 
  
  \hline
  \text{ref+cxt}  &  \textcolor[rgb]{0.40,0.40,0.40}{And through playing music and talking about music, this man }\textcolor[rgb]{0.30,0.30,0.80}{had  }\textcolor[rgb]{0.40,0.40,0.40}{transformed from the paranoid, disturbed man that had just come from walking the streets of downtown Los Angeles to the charming, erudite, brilliant, Juilliard-trained musician. }\text{Music }\textcolor[rgb]{0.90,0.30,0.30}{is }\text{medicine, Music }\textcolor[rgb]{0.90,0.30,0.30}{changes  }\text{us. } \textcolor[rgb]{0.40,0.40,0.40}{And for Nathaniel, music }\textcolor[rgb]{0.30,0.30,0.80}{is  }\textcolor[rgb]{0.40,0.40,0.40}{sanity.} \\
  
  \hline
    \text{sys0}  &  \text{Music }\textcolor[rgb]{0.90,0.30,0.30}{is }\text{medicine, Music }\textcolor[rgb]{0.90,0.30,0.30}{changes  }\text{us. } \textcolor[rgb]{0.40,0.40,0.40}{For nathaniel, music is helping him turn on his mind.} \\
    
    \hline
      \text{sys1}  &  \textcolor[rgb]{0.40,0.40,0.40}{By playing music and talking about music, this person }\textcolor[rgb]{0.30,0.30,0.80}{went  }\textcolor[rgb]{0.40,0.40,0.40}{from being a paranoid, restless, tramp who was just walking on the streets of Los Angeles to a charming, knowledgeable, and outstanding musician who was educated by Juilliard-trained. }\text{Music }\textcolor[rgb]{0.90,0.30,0.30}{was  }\text{medicine . Music }\textcolor[rgb]{0.90,0.30,0.30}{changed  }\text{us . }\textcolor[rgb]{0.40,0.9,0.40}{And  }\textcolor[rgb]{0.40,0.40,0.40}{for nathaniel, music }\textcolor[rgb]{0.30,0.30,0.80}{was  }\textcolor[rgb]{0.40,0.40,0.40}{helping him turn on his mind.} \\
      
    \hline
      \text{sys2}  & \text{Music }\textcolor[rgb]{0.90,0.30,0.30}{is  }\text{medicine, Music is }\textcolor[rgb]{0.90,0.30,0.30}{changes  }\text{us. } \textcolor[rgb]{0.40,0.9,0.40}{And  }\textcolor[rgb]{0.40,0.40,0.40}{for nathaniel, music }\textcolor[rgb]{0.30,0.30,0.80}{is  }\textcolor[rgb]{0.40,0.40,0.40}{helping him turn on the mind .} \\
\hline
\end{tabular}
\caption{\label{table-examples} 
Examples of translation results. sys0: sentence-level transformer. sys1: concatenate baseline. sys2: our context-adaptive DocMT. A unified grammatical tense sometimes reduces the quality of translation.
}
\end{table*}

This is an appendix.

\end{CJK}  
\end{document}